\documentclass[letterpaper]{article} 
\usepackage{aaai2026}  
\usepackage{times}  
\usepackage{helvet}  
\usepackage{courier}  
\usepackage[hyphens]{url}  
\usepackage{graphicx} 
\usepackage{natbib}  
\usepackage{caption} 
\usepackage{algorithm}
\usepackage{algorithmic}
\usepackage{lineno}
\usepackage{amsmath} 
\usepackage{amssymb}
\usepackage{float}    
\usepackage{colortbl}
\usepackage{xcolor}
\usepackage{multirow}
\usepackage{cleveref}
\usepackage{booktabs}
\usepackage{pifont}

\definecolor{firstColor}{rgb}{1.0, 0.7098, 0.6392}
\definecolor{secondColor}{rgb}{1.0, 0.8627, 0.7843}
\definecolor{thirdColor}{rgb}{1.0, 1.0, 0.7843}

\newcommand{\Skip}[1] {}
\newcommand{\padm}{\emph{Pose-Aware Diffusion Model}}
\newcommand{\pf}{\textbf{PFAvatar}}

\usepackage{newfloat}
\usepackage{listings}
\DeclareCaptionStyle{ruled}{labelfont=normalfont,labelsep=colon,strut=off} 
\floatstyle{ruled}
\newfloat{listing}{tb}{lst}{}
\floatname{listing}{Listing}
\title{PFAvatar: Pose-Fusion 3D Personalized Avatar Reconstruction from Real-World Outfit-of-the-Day Photos}
\author{
    Dianbing Xi\textsuperscript{\rm 1, \footnotemark[1]},
    Guoyuan An\textsuperscript{\rm 2, \footnotemark[1]},
    Jingsen Zhu\textsuperscript{\rm 3, \thanks{These authors contributed equally.}},
    Zhijian Liu\textsuperscript{\rm 1},
    Yuan Liu\textsuperscript{\rm 4},
    Ruiyuan Zhang\textsuperscript{\rm 5},
    Jiayuan Lu\textsuperscript{\rm 1},
    Yuchi Huo\textsuperscript{\rm 1, \thanks{These authors are the corresponding authors.}},
    Rui Wang\textsuperscript{\rm 1}
}
\affiliations{
    \textsuperscript{\rm 1}State Key Laboratory of CAD\&CG, Zhejiang University\\
    \textsuperscript{\rm 2}Independent Contributor\\
    \textsuperscript{\rm 3}Cornell University\\
    \textsuperscript{\rm 4}Hong Kong University of Science and Technology\\
    \textsuperscript{\rm 5}Huawei Technologies Ltd\\

}

\usepackage{bibentry}

\begin{document}

\maketitle

\begin{abstract}

We propose PFAvatar (Pose-Fusion Avatar), a new method that reconstructs high-quality 3D avatars from ``Outfit of the Day'' (OOTD) photos, which exhibit diverse poses, occlusions, and complex backgrounds. Our method consists of two stages: (1) fine-tuning a pose-aware diffusion model from few-shot OOTD examples and (2) distilling a 3D avatar represented by a neural radiance field (NeRF). In the first stage, unlike previous methods that segment images into assets (e.g., garments, accessories) for 3D assembly, which is prone to inconsistency, we avoid decomposition and directly model the full-body appearance. By integrating a pre-trained ControlNet for pose estimation and a novel Condition Prior Preservation Loss (CPPL), our method enables end-to-end learning of fine details while mitigating language drift in few-shot training. Our method completes personalization in just 5 minutes, achieving a 48$\times$ speed-up compared to previous approaches. In the second stage, we introduce a NeRF-based avatar representation optimized by canonical SMPL-X space sampling and Multi-Resolution 3D-SDS. Compared to mesh-based representations that suffer from resolution-dependent discretization and erroneous occluded geometry, our continuous radiance field can preserve high-frequency textures (e.g., hair) and handle occlusions correctly through transmittance. 
Experiments demonstrate that PFAvatar outperforms state-of-the-art methods in terms of reconstruction fidelity, detail preservation, and robustness to occlusions/truncations, advancing practical 3D avatar generation from real-world OOTD albums. In addition, the reconstructed 3D avatar supports downstream applications such as virtual try-on, animation, and human video reenactment, further demonstrating the versatility and practical value of our approach.
\end{abstract}
  
\section{Introduction}
\label{sec:intro}

This paper focuses on a novel and essential task: transforming everyday photo albums into textured, personalized 3D human models~\cite{xiu2024puzzleavatarassembling3davatars}. These photo collections, commonly referred to as “Outfit of the Day” (OOTD) photos, exhibit several defining characteristics: 1) consistent identity, outfit, hairstyle, and accessories across images, 2) diverse poses and scales, 3) frequent occlusions and significant truncations, and 4) varied viewpoints against complex backgrounds. These characteristics pose significant challenges to existing 3D avatar reconstruction methods~\cite{5473199,10.1145/3072959.3083722,xiong2024mvhumannet,shen2023xavatarexpressivehumanavatars,I_k_2023}, which typically require full visibility of the subject and precise camera calibration. 

To address these challenges, PuzzleAvatar~\cite{xiu2024puzzleavatarassembling3davatars} proposes an innovative framework that eliminates the need for camera calibration. It first \emph{decomposes} OOTD photos into multiple semantic assets (e.g., garments, accessories, faces, and hair), each of which is associated with a unique token in a Stable Diffusion~\cite{rombach2021highresolution} model—a component referred to as {PuzzleBooth}. After fine-tuning the diffusion model to personalize it for the given assets, the learned tokens are treated as ``puzzle pieces'' and used to assemble a 3D personalized avatar through a mesh-based representation using DMTet~\cite{gao2020learningdeformabletetrahedralmeshes}.

However, this method faces several key challenges. First, although decomposing photos into multiple assets helps mitigate issues such as uncommon body poses, it heavily relies on the accuracy of segmentation, which may introduce visual inconsistencies. Second, since PuzzleBooth assembles the human body from individual pieces, it does not support pose-controllable image generation. Moreover, learning multiple individual components significantly increases the overall training time, making the process inefficient.
As a result, applying Score Distillation Sampling (SDS) \cite{wang2022scorejacobianchaininglifting,poole2022dreamfusiontextto3dusing2d} for 3D reconstruction often leads to the Janus problem due to inconsistent pose alignment. Finally, the topology of DMTet is restricted by the initial mesh structure, making it difficult to represent complex topological variations. Given the highly diverse appearances in OOTD images, this limitation often results in suboptimal 3D geometry with poor visual fidelity (e.g., hair strands or clothing textures).

This paper presents a novel approach, \textbf{PFAvatar}, to enhance the visual quality of 3D avatars. First, unlike methods that decompose OOTD images into multiple assets, we propose an end-to-end pose-aware diffusion model. We utilize a pre-trained ControlNet \cite{zhang2023addingconditionalcontroltexttoimage} to predict the pose of each image, providing pose-prior information for training the personalized diffusion model.

\twocolumn[{%
\renewcommand\twocolumn[1][]{#1}%
\maketitle
\begin{center}
\centering
\includegraphics[width=\linewidth]{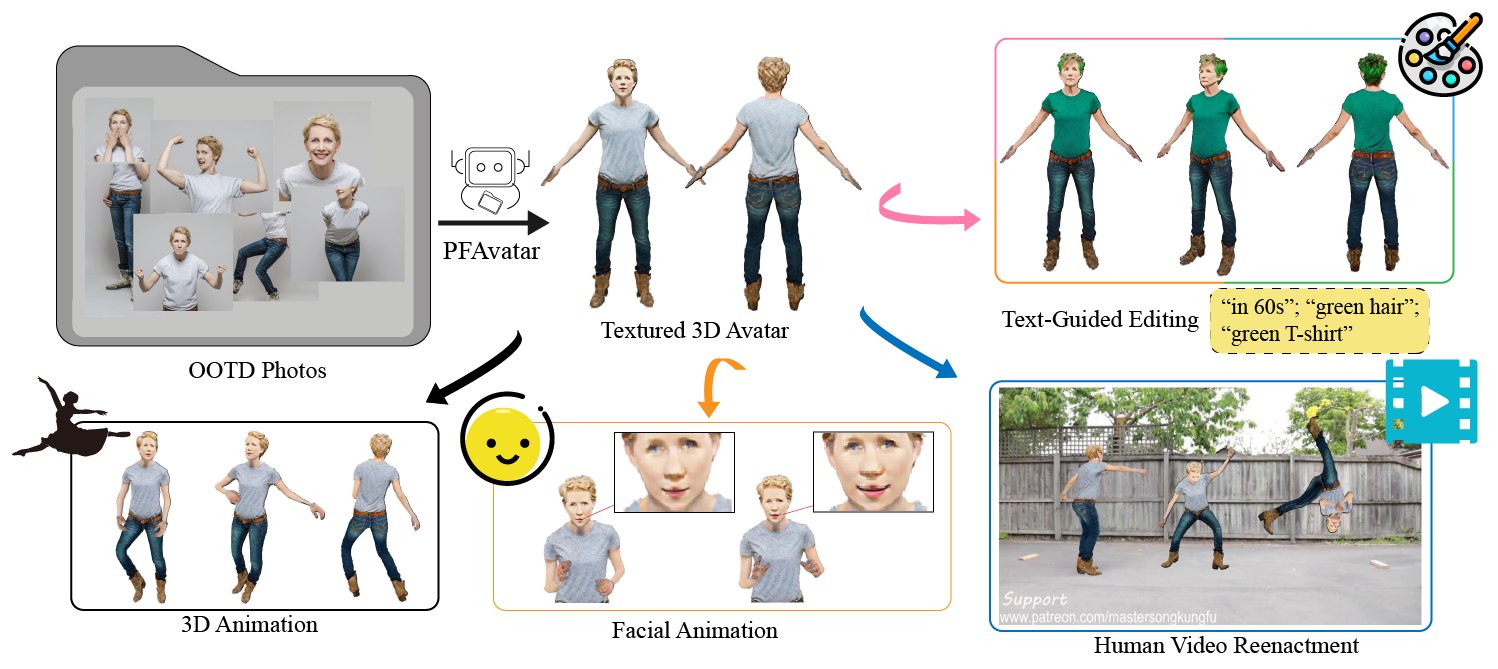}
\captionof{figure}{Using the ``Outfit of the Day'' (OOTD) photos from a personal collection (shown in the upper left), our PFAvatar reconstructs a personalized and fully textured 3D NeRF avatar (depicted in the middle). These OOTD photos can vary widely in terms of body poses, scales, camera angles, framing, frequent partial occlusions, or significant truncation. PFAvatar is designed to handle such variability robustly, enabling a range of downstream tasks. These include virtual try-on through text-guided editing, 3D animation, facial animation, and human video reenactment, all while meticulously preserving the subject's identity and unique characteristics.}
\label{fig:teaser}
\end{center}%
}]

Additionally, we introduce a novel Condition Prior Preservation Loss (CPPL) to mitigate language and control drift issues that often arise during few-shot data finetuning \cite{ruiz2023dreamboothfinetuningtexttoimage}. By eliminating asset decomposition, our method enables end-to-end learning of detailed structures, significantly improving both consistency and accuracy in the final output.

Second, we distill a Neural Radiance Field (NeRF) representation instead of a mesh from our trained diffusion model as the 3D avatar. Our motivation is that NeRF offers greater flexibility and robustness for this task than mesh-based representations. Specifically, NeRF’s volume density naturally handles occlusions through transmittance, ensuring that rays intersecting occluders contribute less to the final pixel and avoiding the generation of false surfaces. Additionally, 
Compared to mesh-based approach that struggles with high-frequency details due to resolution-dependent discretization, NeRF’s continuous volume rendering directly integrates high-frequency positional embeddings (e.g., hashgrids~\cite{mueller2022instant}) into the radiance field, which preserves fine-grained details, such as hair strands and intricate patterns, without loss of fidelity.

We conducted extensive explorations to identify the most effective method for distilling high-quality NeRF representations from the fine-tuned Pose-Aware Diffusion Model. Our final approach involves sampling from both the canonical SMPL-X space and the observation space. Sampling from the canonical space generates additional pose-conditioned images, which ensures 3D-consistent NeRF optimization. For the observation space, we employ a progressive sampling method~\cite{huang2023dreamwaltzmakescenecomplex} to achieve higher-quality appearance details.

Additionally, we observed the instability issue of Score Distillation Sampling (SDS) \cite{poole2022dreamfusiontextto3dusing2d,lin2023magic3dhighresolutiontextto3dcontent}, as highlighted in \cite{cao2024dreamavatar}. This instability degrades our fine local structures due to a lack of human priors. To address this, we introduce a Local Geometry Loss during NeRF training. This loss leverages predefined meshes of body parts, such as hands and faces, to preserve intricate details and enhance the overall fidelity of the 3D avatar.
Although NeRF representation is relatively new and lacks the extensive manipulation tools available for traditional mesh-based methods, we demonstrate that our NeRF-based 3D avatar is capable of animation and video reenactment. Moreover, it shows good imaging quality and detail preservation, showcasing its potential as a superior alternative for 3D avatar generation.
Extensive experiments on real-world datasets, as shown in Fig.~\ref{fig:qualitative-1} and Fig.~\ref{fig:qualitative-2}, demonstrate that {PFAvatar} outperforms state-of-the-art methods in avatar reconstruction and editing from OOTD photos. We also conducted a user study (see Appendix), which found that users prefer our model over state-of-the-art 3D avatar techniques, further validating the effectiveness and appeal of our approach.
In summary, our main contributions include:

\begin{itemize}
    \item We propose a {Pose-Aware Diffusion Model} trained from OOTD photos. By leveraging a pre-trained ControlNet for pose estimation and a Condition Prior Preservation Loss (CPPL), our method eliminates reliance on decomposition and enables end-to-end learning of detailed structures with minimal language/control drift under few-shot settings.

    \item We distill a personalized NeRF avatar from this diffusion model using canonical SMPL-X sampling for 3D consistency and Multi-Resolution 3D-SDS for high-fidelity details. To address SDS instability, we introduce a Local Geometry Loss on anatomical part meshes, preserving fine structures such as hands and faces.

    \item Extensive experiments show that our NeRF-based avatars outperform mesh-based baselines in reconstruction quality, detail preservation, and robustness to occlusions and truncations.
\end{itemize}

\section{Related Work}
\subsection{Text and Image-guided 3D Avatar Generation} 
Numerous studies have explored reconstructing clothed humans from multi-view images \cite{lin2024fasthuman,saito2019pifupixelalignedimplicitfunction,peng2021neuralbodyimplicitneural} or monocular video \cite{weng_humannerf_2022_cvpr,li2020monocularrealtimevolumetricperformance}. Recent work leverages text-to-image models and SDS to generate avatars from language descriptions, achieving finer details \cite{wang2023disentangledclothedavatargeneration,liao2023tadatextanimatabledigital,kolotouros2023dreamhumananimatable3davatars,huang2023dreamwaltzmakescenecomplex,cao2023dreamavatartextandshapeguided3d}. 
In addition, methods such as~\cite{li2024pshuman,qiu2025LHM,pan2024humansplat} typically depend on accurate pose estimation and require fully visible bodies, making them incapable of reconstructing avatars from images with significant truncations. In contrast, {ours} overcomes these limitations, enabling robust 3D avatar reconstruction from unconstrained, everyday photos without strict visibility constraints.

{
\begin{figure*}[ht]
\includegraphics[scale=0.13]{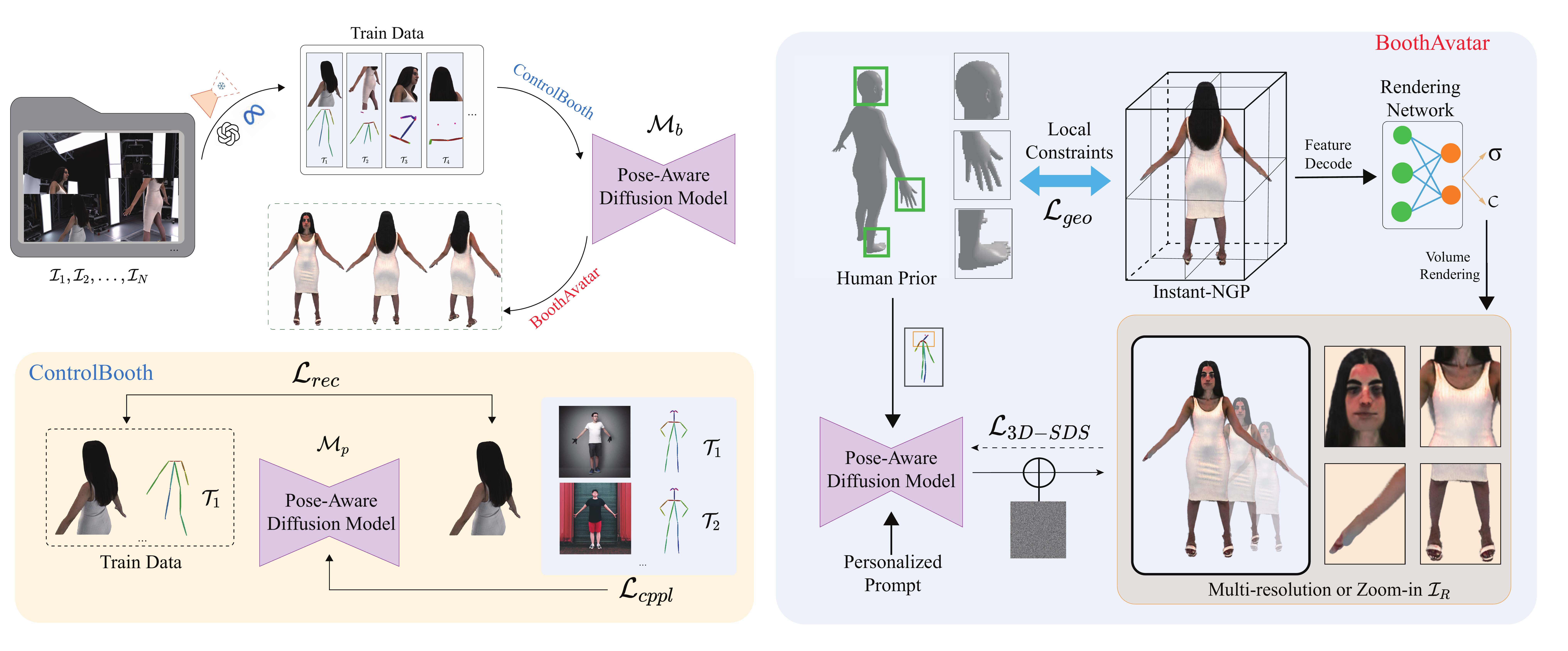}
  \caption{\textbf{Overview of our \pf{}  pipeline.} 
    (1) The top left illustrates the overall flow of our framework, which consists of two main stages: ControlBooth and BoothAvatar.
    (2) The bottom left displays the training details of the ControlBooth stage. In this stage, our input data is composed of three parts: images, pose-conditioning, and captions. These are used to fine-tune a Pose-Aware Diffusion Model $\mathcal{M}_\text{b}$, where the Text-Encoder and the UNET are trained using the reconstruction diffusion loss, $\mathcal{L}_{\text{rec}}$(\cref{eq:controlbooth_rec}), and the condition-based prior preservation loss, $\mathcal{L}_{\text{cppl}}$(\cref{eq:controlbooth_cppl}).
    (3) The right section shows the details of the BoothAvatar stage. In this stage, the avatar is represented as an A-posed canonical avatar. The model $\mathcal{M}_\text{b}$ obtained from the previous stage is used to guide this reconstruction process. Using multi-resolution $\mathcal{L}_{\text{3D-SDS}}$ (\cref{eq:3d-sds}), we optimize a NeRF represented by Instant-NGP, with an additional loss $\mathcal{L}_{\text{geo}}$ (\cref{eq:geo}) to stabilize local structures during SDS optimization.}
    \label{fig:overview}
\end{figure*}
}

\subsection{Finetuning of Personalized Diffusion Models}
In recent years, with the increasing interest in the text-to-image domain, pioneering researchers have begun exploring methods for personalizing text-to-image models using photos of specific subjects. Work on model customization introduces new concepts through fine-tuning (either partial or whole) of pre-trained networks\cite{avrahami2023bas,jain2022zeroshottextguidedobjectgeneration,kumari2023multiconceptcustomizationtexttoimagediffusion,liu2024parameterefficientorthogonalfinetuningbutterfly,ruiz2023dreamboothfinetuningtexttoimage}. Other research re-purposes diffusion models for new tasks \cite{fu2024geowizardunleashingdiffusionpriors,ke2024repurposingdiffusionbasedimagegenerators,kocsis2024intrinsicimagediffusionindoor}. One representative work is DreamBooth\cite{ruiz2023dreamboothfinetuningtexttoimage}, which uses a rare token to represent a specific subject or style, while preventing overfitting through a prior preservation loss. Another approach, textual inversion \cite{gal2022textual}, generates a new embedding for the input concept and optimizes this embedding vector with a few photos to enable subject-driven image generation. 

Despite these methods achieving laudable results with common objects, the abundance of prior information in the human body poses challenges. This hinders the incorporation of such prior control when fine-tuning human images. Consequently, consistency may diminish when integrating with controllers like ControlNet\cite{zhang2023addingconditionalcontroltexttoimage}.

\subsection{Pose-Free Reconstruction in the Wild}

In our context, pose includes both camera position and body articulation. Accurate camera pose is critical for 3D reconstruction~\cite{mildenhall2021nerf}, but estimating it from unconstrained OOTD photos is highly challenging.

Prior works address this by jointly optimizing object and camera parameters~\cite{Xia_2022_BMVC,wang2021nerfmm,lin2021barf}, leveraging geometric priors~\cite{meuleman2023localrf,fu2023colmapfree,bian2022nopenerf}, or using learning-based methods~\cite{zhang2024camerasraysposeestimation,wang2023dust3rgeometric3dvision,wang2023pflrmposefreelargereconstruction}. Body pose estimation is even harder due to higher dimensionality, and most static-scene methods~\cite{Sun_2022,martinbrualla2021nerfwildneuralradiance} are not suitable for articulated subjects like humans.

PuzzleAvatar~\cite{xiu2024puzzleavatarassembling3davatars} and AvatarBooth~\cite{zeng2023avatarboothhighqualitycustomizable3d} generate animatable avatars from a few images. However, AvatarBooth relies on dual diffusion models and SDS~\cite{poole2022dreamfusiontextto3dusing2d}, while PuzzleAvatar, based on Break-A-Scene~\cite{avrahami2023bas}, must learn multiple concepts and requires ~4 hours of training. 

In contrast, our {PFAvatar} completes personalization in just 5 minutes. Unlike NFSD~\cite{katzir2023noisefreescoredistillation} used by PuzzleAvatar, our method incorporates human-specific priors, reducing Janus artifacts and improving reconstruction fidelity.

\section{Method}
\label{sec:method}
Our PFAvatar reconstructs a 3D avatar that faithfully captures both geometry and appearance from input OOTD Photos $\{ \mathcal{I}_i \}$.
As shown in Fig. \ref{fig:overview}, it is composed of two primary stages. The first stage, \textbf{ControlBooth} , pre-processes the input images and uses them to fine-tune a Pose-Aware Diffusion Model $\mathcal{M}_\text{b}$. The second stage, \textbf{BoothAvatar}, distills a NeRF-based 3D avatar model from $\mathcal{M}_\text{b}$.

\subsection{ControlBooth: Injecting Avatar Feature to Pose-Aware Diffusion Model}
\label{sec:controlbooth}
In this stage, we aim to train a personalized diffusion model $\mathcal{M}_\text{b}$ to achieve novel view synthesis results based on the OOTD photos.

\paragraph{Pose-Aware Diffusion Model:} OOTD images often exhibit diverse poses and frequent occlusions, making it challenging to train diffusion models directly on such data. Previous approaches, such as PuzzleAvatar~\cite{xiu2024puzzleavatarassembling3davatars}, address this by segmenting OOTD images into multiple assets (e.g., garments, accessories, faces, and hair), each associated with a unique token in a Stable Diffusion model~\cite{rombach2021highresolution}. However, this method can introduce segmentation inconsistencies, such as mismatched segment boundaries or incorrect labeling of parts.

To circumvent these issues, we propose a {{Pose-Aware Diffusion Model}} conditioned on input images $\{ \mathcal{I}_i \}$, poses $\{ \mathcal{P}_i \}$, and text prompts $\{ \mathcal{T}_i \}$. Our data preprocessing pipeline begins by using Ground-SAM~\cite{ren2024grounded} to separate the foreground (avatar) from the background, yielding the input images $\{ \mathcal{I}_i \}$. Importantly, this process only isolates the human region from the background, avoiding the segmentation inconsistency problems inherent in fine-grained part-based segmentation. Next, we employ a pre-trained ControlNet\cite{zhang2023addingconditionalcontroltexttoimage} to predict the poses $\{ \mathcal{P}_i \}$ for all images. Finally, we utilize a vision-language model (GPT-4V~\cite{GPT-4V}) to generate detailed textual descriptions $\mathcal{T}_i$ for each image. These descriptions are obtained through a specially designed query prompt (see Appendix), capturing fine-grained attributes such as body orientation, hairstyle, clothing, and other relevant characteristics.

\paragraph{Finetuning Pose-Aware Diffusion Model:} Our diffusion model is finetuned on the inputs described above using two loss terms: a {Reconstruction Diffusion Loss} $\mathcal{L}_{\text{rec}}$ 
to improve subject fidelity and a novel Condition Preservation Prior Loss (CPPL) $\mathcal{L}_{\text{cppl}}$ to prevent the degradation of controllability. The overall training objective is defined as $\mathcal{L}_{\text{total}}^\text{CB} = \mathcal{L}_{\text{rec}} + \lambda_{\text{cppl}}\mathcal{L}_{\text{cppl}}$, where $\lambda_{\text{cppl}} = 1$.

The $\mathcal{L}_{\text{rec}}$ encourages fidelity for each data tuple of prompt \( \mathcal{T}_i \), pose \( \mathcal{P}_i \), and image \( \mathcal{I}_i \) by ensuring that the model maintains high fidelity to the input description. For simplicity, we omit $\epsilon$, $t$, and $w_t$ in all subsequent loss equations. This loss is defined as follows:
\begin{equation}
    \label{eq:controlbooth_rec}
    \mathcal{L}_{\text{rec}} = \mathbb{E} \left[\left\| \mathcal{D}_{\theta}(\alpha_t \mathcal{I}_{i} + \sigma_t \epsilon, \mathbf{c}_{t_i}, \mathbf{c}_{p_i}) - \mathcal{I}_{i} \right\|_2^2 \right],
\end{equation}
where the $i$-th image-space condition \( \mathcal{P}_i \) encodes a feature space conditioning vector \( \mathbf{c}_{p_i} \), and \( \mathbf{c}_{t_i} \) represents its corresponding text conditioning vector.

\begin{figure}
  \centering
   \includegraphics[width=1.0\linewidth]{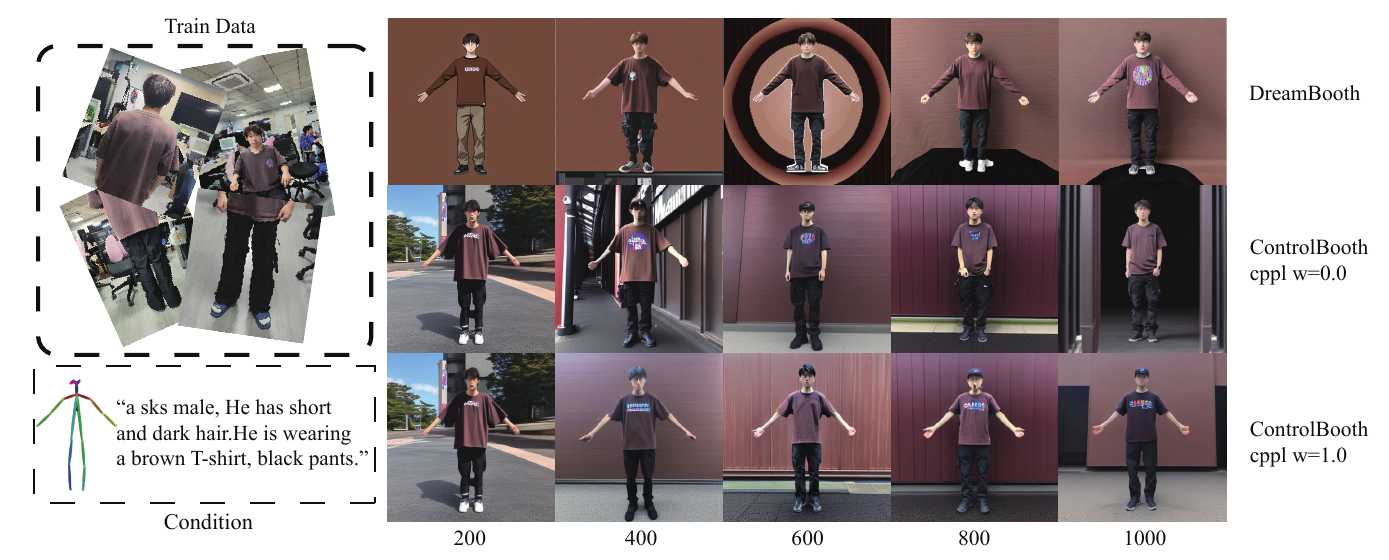}
    \caption{\textbf{Balancing diversity and control with Condition Prior-Preservation Loss (CPPL).} Using the fine-tuning strategy of Naive DreamBooth (Row 1) to generate images with new poses may introduce color discrepancies, significantly reducing thematic consistency. Training with only \cref{eq:controlbooth_rec} may lead to overfitting on the context of the input image and the subject's appearance (e.g., pose). CPPL (Row 3) acts as a regularizer, mitigating overfitting while encouraging diversity and maintaining control.}
    \label{fig:cppl}
\end{figure}

When finetuning on a small set of OOTD images, there is a risk of reducing the variability in the output poses and view elements of the avatar (e.g., snapping to the few-shot views), as shown in Fig.\ref{fig:cppl} (middle row). To address these issues, we propose a novel \textbf{{Condition Prior Preservation Loss (CPPL)}} to prevent the degradation of controllability, which promotes diversity, counters language drift, and helps maintain control capabilities.

Specifically, we generate prior preservation (pr) data \( \mathcal{I}_{pr} = \mathcal{D}_{\theta}(\epsilon, \mathbf{c}_{prt}, \mathbf{c}_{prp}) \) using the ancestral sampler on the frozen pre-trained T2I diffusion model with random initial noise \( \epsilon \sim \mathcal{N}(0, 1) \) and prior preservation text (prt) conditioning vectors \( \mathbf{c}_{prt} := \Gamma(f_t(\mathcal{T}_{pr})) \) and prior preservation pose (prp) condition vectors \( \mathbf{c}_{prp} := \mathcal{F}(\mathcal{P}_{pr}) \).
Here, \( f_t \) is used to convert the prompt \( \mathcal{T}_{pr} \) into the corresponding text embedding, while \( \Gamma \) represents the text encoder that transforms it into the corresponding text conditioning vectors. \( \mathcal{F} \) is a neural block that converts the output 2D map \( \mathcal{P}_{pr} \) into \( \mathbf{c}_{prp} \). The form of \( \mathcal{L}_\text{cppl} \) is given by:

\begin{equation}
\label{eq:controlbooth_cppl}
\mathbb{E} \left[ \lambda w_t' \left\| \mathcal{D}_{\theta}(\alpha_t \mathcal{I}_{pr_i} + \sigma_t \epsilon, \mathbf{c}_{prt_i}, \mathbf{c}_{prp_i}) - \mathcal{I}_{pr_i} \right\|_2^2 \right],    
\end{equation}

where \( \mathbf{c}_{prp_i}, \mathbf{c}_{prt_i}, \mathcal{I}_{pr_i} \) are simplified as \( \mathbf{c}_{p_i}, \mathbf{c}_{t_i}, \mathcal{I}_{i} \). \( \mathcal{L}_\text{cppl} \) is a condition prior-preservation term that supervises the model with its own generated images.

\subsection{BoothAvatar: Reconstructing Avatar from the Finetuned Model}
\label{sec:boothavatar}

In this stage, we reconstruct a NeRF-based \cite{mildenhall2021nerf} canonical A-posed subject-consistent avatar based on the Pose-aware diffusion model $\mathcal{M}_\text{b}$ obtained from the previous stage. To fully leverage the controllability of $\mathcal{M}_\text{b}$, we employ the following techniques to reconstruct a high-quality 3D avatar.

\paragraph{NeRF-Based 3D Representation:} Neural Radiance Fields (NeRF) \cite{mildenhall2021nerf} have shown more flexible 3D representation capabilities compared to mesh-based representations (e.g., PuzzleAvatar's DMTet~\cite{gao2020learningdeformabletetrahedralmeshes}), which is more effective for reconstructing complex clothed human avatars. We use Instant-NGP \cite{mueller2022instant} as our canonical avatar representation, which has a high training speed while maintaining high-quality representation. We train the NeRF-based avatar model using score distillation sampling (SDS)~\cite{poole2022dreamfusiontextto3dusing2d} from controllable novel views generated by our personalized model $\mathcal{M}_{b}$.

Specifically, we sample camera poses from the observation space and sample avatar poses from the canonical SMPL-X space. An additional advantage of sampling from the canonical SMPL-X space is its capability to generate skeleton images from the current viewpoint, thereby facilitating convergence and ensuring thematic consistency.

\paragraph{3D-Consistent Score Distillation Sampling:} 
Pose-aware 3D-consistent SDS sampling techniques allow for more controllable processes with high thematic consistency. Previous works that relied solely on text-guided SDS often faced Janus artifact issues \cite{xiu2024puzzleavatarassembling3davatars, zeng2023avatarboothhighqualitycustomizable3d}. To address this, we incorporate 3D-aware conditioning images to refine SDS \cite{huang2023dreamwaltzmakescenecomplex} to facilitate 3D-consistent NeRF optimization. Specifically, an extra conditioning image \(c\) is used to compute the 3D-SDS gradient $\nabla_{\boldsymbol{\theta}} \mathcal{L}_{\mathrm{3D-SDS}}(\phi, \mathbf{x})$ as follows:
{
\begin{equation}\label{eq:3d-sds}
\mathbb{E}\left[w(t)\left(\boldsymbol{\epsilon}_{\phi}\left(\mathbf{x}_{t} ; y, t, c\right) - \boldsymbol{\epsilon}\right) \frac{\partial \mathbf{z}_{t}}{\partial \mathbf{x}} \frac{\partial \mathbf{x}}{\partial \boldsymbol{\theta}}\right],
\end{equation}
}

where the conditioning image \(c\) may consist of one or a combination of elements such as skeletons or depth maps, \(w(t)\) is a weighting function based on the timestep \(t\), and \(y\) represents the corresponding text prompt. In our implementation, we choose skeletons as the conditioning image type due to their minimal structural priors to facilitate complex avatar generation.

\paragraph{Local Geometric Constraint:} The instability of SDS optimization can compromise fine local structures derived from human priors. To address this, we introduce a local geometry loss during NeRF training based on predefined meshes of body parts, such as hands and faces. This loss aligns NeRF densities $\tau$ in local regions with the predefined meshes through a margin ranking loss $\mathcal{L}_{\text{geo}}$:
\begin{equation}
    \label{eq:geo}
    \mathcal{L}_{\text{geo}} = \begin{cases} 
      \left( \max(0, \tau_{\text{max}} - \tau(\mathbf{p})) \right)^2 & \text{if } \mathbf{p} \text{ on mesh}, \\
      \left( \max(0, \tau(\mathbf{p}) - \tau_{\text{min}}) \right)^2 & \text{if } \mathbf{p} \text{ not on mesh},
    \end{cases}
\end{equation}
where $\mathbf{p}$ represents 3D points sampled on or near predefined meshes, $\tau(\mathbf{p})$ denotes the densities of 3D points $\mathbf{p}$ predicted by NeRF, and $\tau_{\text{min}}$, $\tau_{\text{max}}$ are constant hyperparameters.

In summary, at the BoothAvatar stage, the total loss to optimize the NeRF-based avatar representation is:
\begin{equation}
    \mathcal{L}_{\text{total}}^{\text{BA}} = \mathcal{L}_{\text{3D-SDS}} + \lambda_{\text{geo}} \mathcal{L}_{\text{geo}},\tag{5}
\end{equation}
where $\lambda_{\text{geo}} = 1.0$.

\paragraph{Multi-Resolution and Zoom-in Sampling:} Directly rendering high-resolution images from NeRFs is computationally expensive. A common approach is to render a low-resolution image and then up-sample it to a higher resolution for SDS training \cite{chen2023fantasia3ddisentanglinggeometryappearance,lin2023magic3dhighresolutiontextto3dcontent}. However, directly increasing the upsampled resolution can lead to training collapse or inconsistent appearances. To address this, we leverage a multi-resolution and zoom-in optimization strategy, which progressively increases the up-sampling resolution for more stable SDS training while refining the avatar appearance.

\begin{figure}[ht]
\centering
\includegraphics[width=1.0\linewidth]{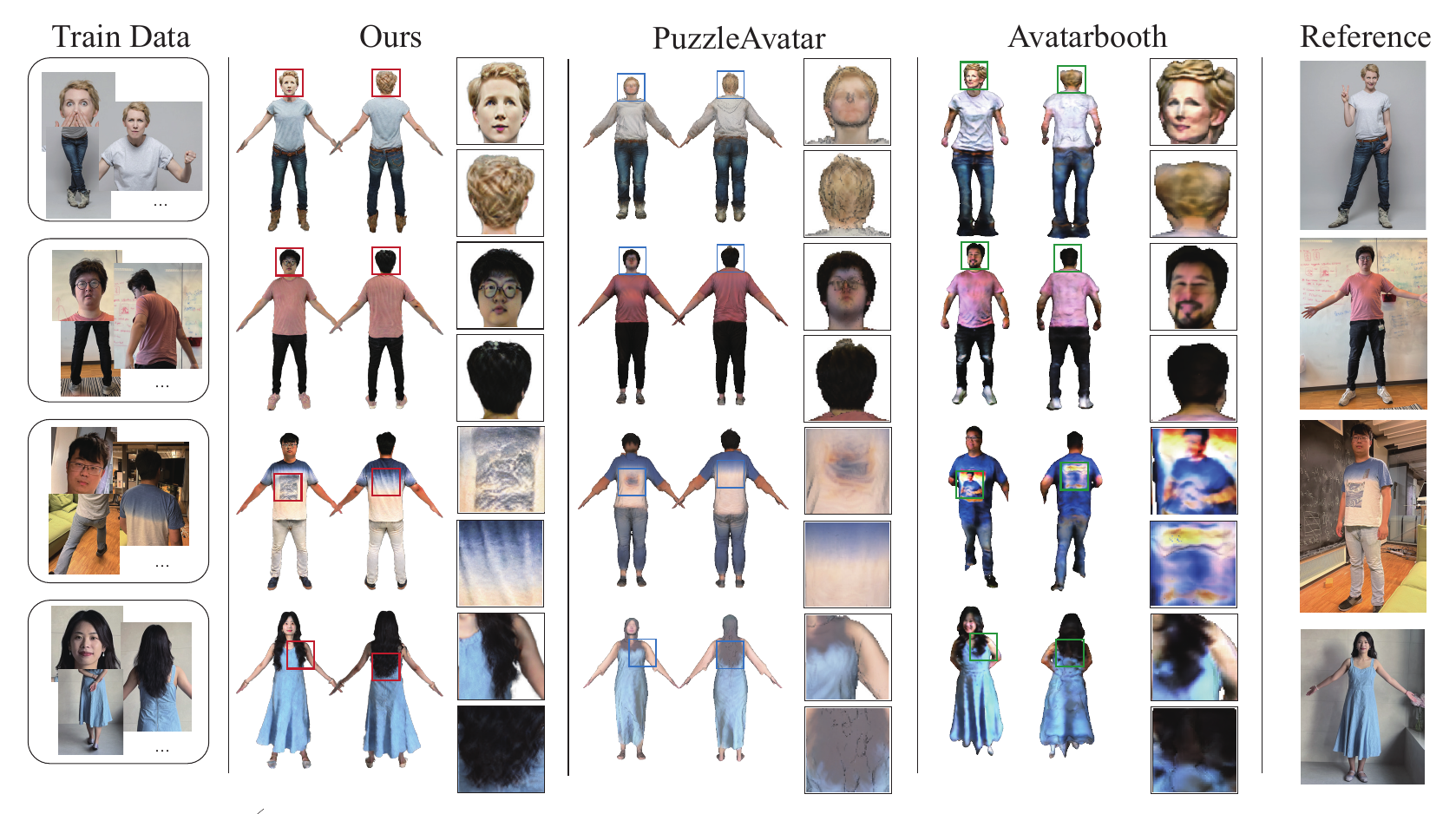}
    \caption{\textbf{Qualitative Comparison I: Custom Dataset.} Visual results on three distinct subjects compare PuzzleAvatar, AvatarBooth, and our method (PFAvatar). Our approach consistently preserves finer details and maintains structural coherence across multiple views, demonstrating clear superiority over the baselines. This highlights PFAvatar's robustness in achieving high-quality, consistent reconstructions under challenging conditions.}
    \label{fig:qualitative-1}
\end{figure}

\section{Experiment}

\subsection{Implementation Details}
\paragraph{Dataset.} We evaluate our system's performance using two distinct datasets: the {Custom} dataset and the {PuzzleIOI} dataset~\cite{xiu2024puzzleavatarassembling3davatars}, which serves as a standardized benchmark for personalized human avatar reconstruction. In addition, we introduce a {Custom} dataset to assess the model's performance on real-world, in-the-wild scenarios. This dataset comprises casual OOTD photos captured using mobile devices under unconstrained conditions. The images feature diverse poses, complex backgrounds, and varied camera viewpoints, posing significant challenges for 3D reconstruction. This setting enables a robust evaluation of the model’s generalization ability in realistic environments.

\paragraph{Baselines.} 
To evaluate pose-aware diffusion model identity preservation, we compare our method with state-of-the-art (SOTA) approaches, including PuzzleAvatar \cite{xiu2024puzzleavatarassembling3davatars}, InstantID \cite{wang2024instantid}, FreeCustom \cite{ding2024freecustom}, and DisenBooth \cite{chen2023disenbooth}. For reconstruction quality, we benchmark our method against PuzzleAvatar \cite{xiu2024puzzleavatarassembling3davatars} and AvatarBooth \cite{avrahami2023bas}, both of which are designed to handle unconstrained inputs.

\paragraph{Metrics.}

One important aspect to evaluate is
subject fidelity: the preservation of subject details in generated images. Fellow \cite{ruiz2023dreamboothfinetuningtexttoimage},\cite{raj2023dreambooth3dsubjectdriventextto3dgeneration}. To evaluate subject identity preservation across the two stages of our pipeline, we adopt three metrics: DINO~\cite{caron2021emergingpropertiesselfsupervisedvision}, CLIP-I~\cite{radford2021learningtransferablevisualmodels}, and CLIP-T~\cite{radford2021learningtransferablevisualmodels}.

In addition, for the PuzzleIOI benchmark, we further evaluate the quality of appearance reconstruction. To this end, we render multi-view color images of the reconstructed avatars and report three standard image quality metrics: PSNR (Peak Signal-to-Noise Ratio), SSIM (Structural Similarity Index), and LPIPS (Learned Perceptual Image Patch Similarity). We highlight the {\colorbox{firstColor}{best}}, {\colorbox{secondColor}{second-best}}, and {\colorbox{thirdColor}{third-best}} results accordingly.

\begin{figure}[ht]
   \centering
\includegraphics[width=\linewidth, trim=0 0 2cm 0, clip]{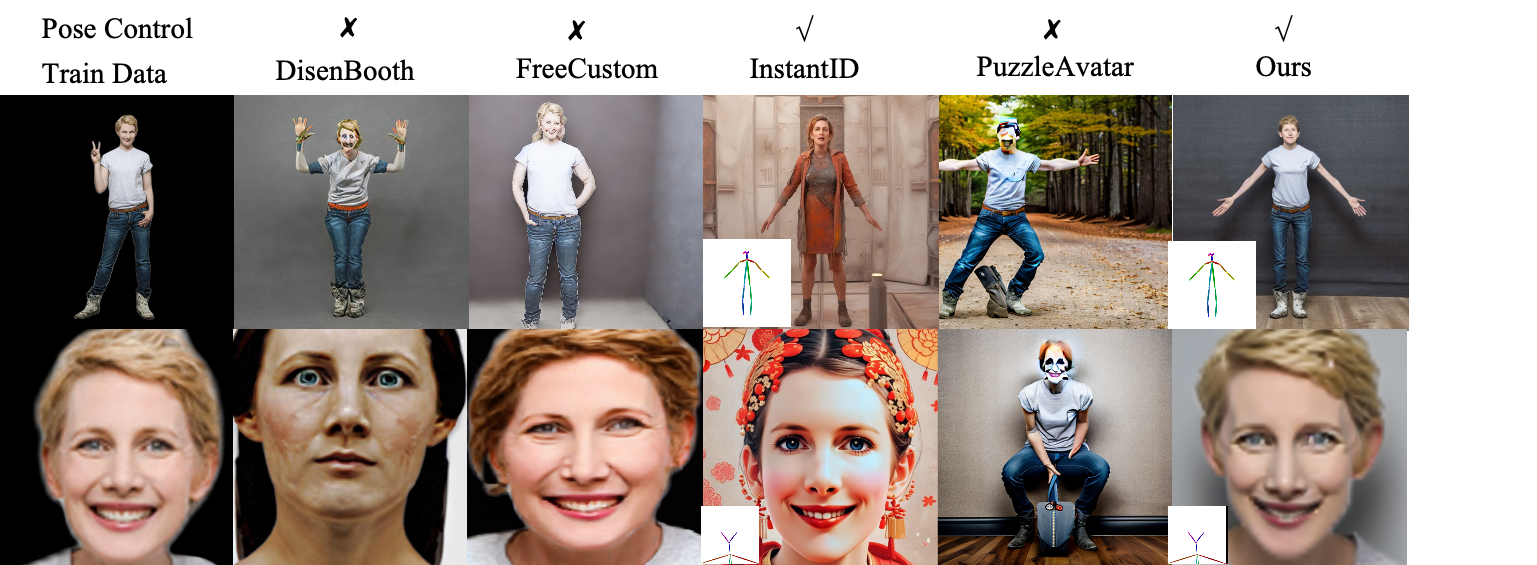}
\caption{\textbf{Qualitative result of personalized diffusion models.} Compared to other methods, our \padm{} achieves superior subject consistency while enabling precise control over character poses. The pose illustration in the lower-left corner represents the input control pose used for conditional generation. \ding{51} indicates support for control pose, while \ding{55} denotes lack of support.}
\label{fig:identiy_compare}
\end{figure}
\begin{table}[ht]
\small
    \centering
    \scalebox{0.8}{
    \begin{tabular}{lcccccc}
        \toprule
        \multirow{2}{*}{\centering Method} & \multicolumn{2}{c}{CLIP-I$\uparrow$} & \multicolumn{2}{c}{DINO$\uparrow$} & \multicolumn{2}{c}{CLIP-T$\uparrow$} \\
        \cmidrule(lr){2-3} \cmidrule(lr){4-5} \cmidrule(lr){6-7}
        & body & head & body & head & body & head \\
        \midrule
        PFAvatar(Ours) & \cellcolor{firstColor}0.9016 & \cellcolor{firstColor}0.9432 & \cellcolor{firstColor} 0.7282 & \cellcolor{firstColor}0.9352 & \cellcolor{firstColor}0.3036 & \cellcolor{firstColor}0.2996 \\
        PuzzleAvatar & 0.8147 & 0.7705 & 0.6257 & 0.6096 & 0.2340 & 0.1849 \\
        InstantID & 0.7687 & \cellcolor{thirdColor}0.8164 & 0.5977 & \cellcolor{thirdColor}0.8302 & 0.2164 & 0.2711 \\
        FreeCustom & \cellcolor{secondColor}0.8573 & \cellcolor{secondColor}0.9337 & \cellcolor{secondColor}0.7022 & \cellcolor{secondColor}0.9222 & \cellcolor{thirdColor}0.2583 & \cellcolor{secondColor}0.2811 \\
        DisenBooth &  \cellcolor{thirdColor}0.8445 & 0.7897 & \cellcolor{thirdColor}0.6930 & 0.8299 & \cellcolor{secondColor}0.2783 & \cellcolor{thirdColor}0.2581 \\
        \bottomrule
    \end{tabular}
    }
    \caption{\textbf{The subject identity preservation performance of different baselines.} Ours achieves the best performance. }
    \label{tab:identity_compare}
\end{table}

\subsection{Experiment Results}
\label{subsec:result}
\paragraph{Analysis of ControlBooth: Subject Identity Preservation}  
As illustrated in Table~\ref{tab:identity_compare} and Fig.~\ref{fig:identiy_compare}, our proposed Pose-Aware Diffusion Model achieves significantly improved subject identity preservation. By leveraging a single diffusion model capable of pose-controllable generation, our method consistently maintains identity fidelity—particularly in the facial and head regions. In contrast, PuzzleAvatar~\cite{xiu2024puzzleavatarassembling3davatars} lacks pose control capability at this stage, leading to suboptimal identity consistency during reconstruction. These results highlight the effectiveness of our strategy, which integrates both pose information and textual descriptions when training on OOTD images, thereby enabling high-fidelity and identity-consistent synthesis.

\begin{figure}
    \centering
    \includegraphics[width=1.0\linewidth]{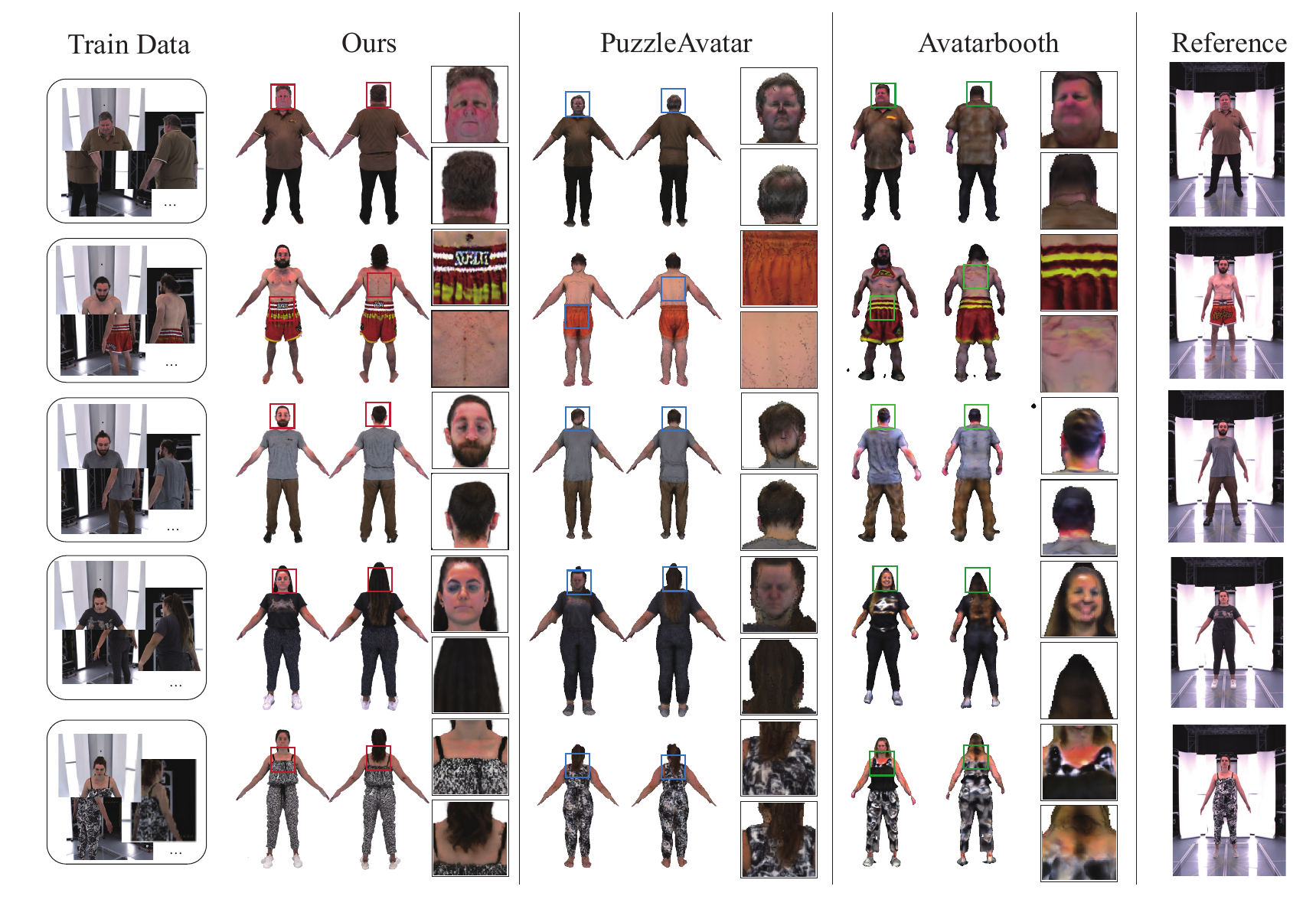}
    \caption{\textbf{Qualitative Comparison-II: PuzzleIOI Dataset.} Visual results on three distinct subjects compare PuzzleAvatar, AvatarBooth, and PFAvatar(Ours). Our approach consistently preserves finer details and maintains structural coherence across multiple views, demonstrating clear superiority over the baselines. This highlights PFAvatar's robustness in achieving high-quality, consistent reconstructions under challenging conditions.}
    \label{fig:qualitative-2}
\end{figure}

\paragraph{Analysis of BoothAvatar: Subject Identity Preservation and Appearance Reconstruction}  
As shown in Table~\ref{tab:comparison_diff_methods}, Fig.~\ref{fig:qualitative-1}, and Fig.~\ref{fig:qualitative-2}, our method achieves superior rendering quality compared to PuzzleAvatar and AvatarBooth, particularly in texture detail preservation, geometric accuracy, and overall visual realism. These improvements demonstrate the robustness and versatility of our approach in generating high-quality 3D avatars from unconstrained, real-world inputs. Moreover, as illustrated in Fig.~\ref{fig:qualitative-2} and Table~\ref{tab:compare_psnr_metric}, our method consistently outperforms PuzzleAvatar~\cite{xiu2024puzzleavatarassembling3davatars}, TeCH~\cite{huang2023techtextguidedreconstructionlifelike}, and AvatarBooth~\cite{zeng2023avatarboothhighqualitycustomizable3d} on the PuzzleIOI benchmark across all evaluation metrics. Specifically, our method produces more realistic textures and achieves more accurate geometric reconstructions—even under challenging conditions such as complex poses—while preserving fine-grained attributes such as facial features, hair strands, and clothing patterns.

\begin{figure}
    \centering
    \includegraphics[width=1.0\linewidth]{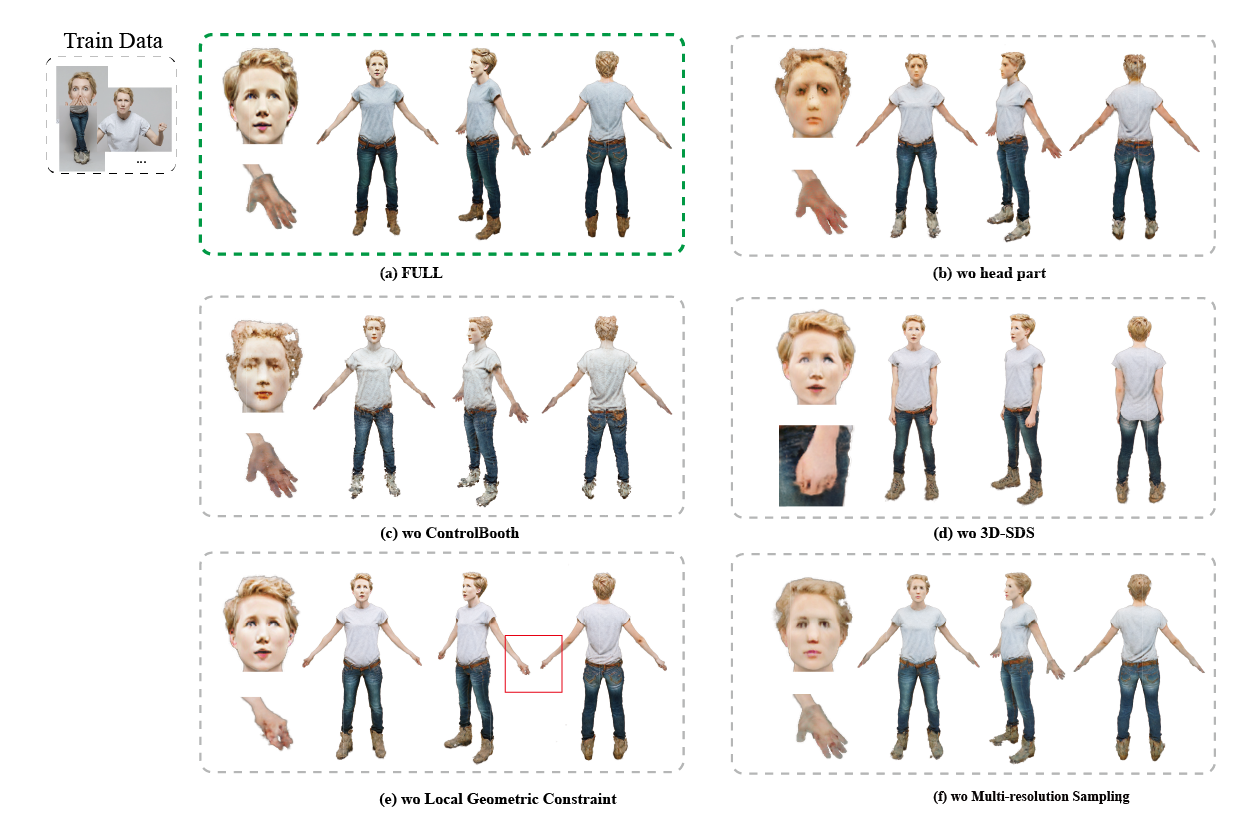}
    \caption{\textbf{Ablation studies.} Qualitative results showing the impact of removing key components. (a) Full model yields the best results; (b–f) each ablation leads to visible degradation in specific aspects.}
    \label{fig:ablation-study}
\end{figure}

\paragraph{Ablation Study} 
We assess the contributions of key components in our framework through ablation experiments. As shown in Fig.~\ref{fig:ablation-study} and summarized in Table~\ref{tab:ablation-study}, each design choice plays a crucial role in overall performance.

\textbf{ControlBooth: Excluding Head Region.}  
To assess the effect of head data, we remove all head samples from training. As shown in Fig.~\ref{fig:ablation-study}~(b), this causes noticeable degradation in reconstruction quality, especially in facial regions.

\textbf{ControlBooth: Using Vanilla DreamBooth.}  
Replacing our customized ControlBooth with DreamBooth~\cite{ruiz2023dreamboothfinetuningtexttoimage} reduces consistency and causes frequent color shifts (e.g., washed-out tones), as shown in Fig.~\ref{fig:ablation-study}~(c), validating the benefit of our targeted training scheme.

\textbf{BoothAvatar: Using Vanilla SDS.}  
Substituting our 3D-SDS with the SDS~\cite{poole2022dreamfusiontextto3dusing2d} compromises the generation of A-pose avatars (Fig.~\ref{fig:ablation-study}~(d)), limiting compatibility with downstream applications. Our 3D-SDS design enables more stable and controllable avatar generation.

\textbf{BoothAvatar: Removing Local Geometric Constraint.}  
Without the local geometric constraint, SDS optimization becomes unstable in fine-grained regions. As illustrated in Fig.~\ref{fig:ablation-study}~(e), this leads to simplified and ambiguous geometry—especially in hands—highlighting the necessity of enforcing localized shape consistency.

\textbf{BoothAvatar: Disabling Multi-Resolution Sampling.}  
Eliminating multi-resolution sampling and training at a fixed resolution results in slower convergence and degraded visual fidelity (Fig.~\ref{fig:ablation-study}~(f)). In contrast, our hierarchical sampling strategy enhances both training efficiency and rendering quality.

\begin{table}[ht]
    \centering
    \scalebox{0.75}{
    \begin{tabular}{lcccccc}
        \toprule
        \multirow{2}{*}{\centering Method} & \multicolumn{2}{c}{CLIP-I$\uparrow$} & \multicolumn{2}{c}{DINO$\uparrow$} & \multicolumn{2}{c}{CLIP-T$\uparrow$} \\
        \cmidrule(lr){2-3} \cmidrule(lr){4-5} \cmidrule(lr){6-7}
        & body & head & body & head & body & head \\
        \midrule
        PFAvatar(Ours) & \cellcolor{firstColor}{0.9125} & \cellcolor{firstColor}{0.9042} & \cellcolor{firstColor}{0.8072} & \cellcolor{firstColor}{0.8617} & \cellcolor{firstColor}{0.3546} & \cellcolor{firstColor}{0.3124} \\
        PuzzleAvatar & \cellcolor{secondColor}{0.8722} & \cellcolor{thirdColor}{0.8652} & \cellcolor{secondColor}{0.722} & \cellcolor{secondColor}{0.8017} & \cellcolor{thirdColor}{0.2836} & \cellcolor{secondColor}{0.2823} \\
        Avatarbooth & \cellcolor{thirdColor}0.8533 & \cellcolor{secondColor}0.8837 & \cellcolor{thirdColor}0.6778 & \cellcolor{thirdColor}0.7869 & \cellcolor{secondColor}0.2907 & \cellcolor{thirdColor}0.2581 \\
        \bottomrule
    \end{tabular}
    }
\caption{\textbf{Qualitative comparison on subject identity preservation.} 
Final rendered avatars from {PFAvatar}, {PuzzleAvatar}, and {AvatarBooth} are compared. 
 Ours demonstrates superior preservation of subject identity in both body and head regions.}
    \label{tab:comparison_diff_methods}
\end{table}
\begin{table}[h!]
\centering
\begin{tabular}{lccc}
\hline
{Method} & {PSNR}$\uparrow$ & {SSIM}$\uparrow$ & {LPIPS}$\downarrow$ \\
\hline
PFAvatar(Ours)       & \cellcolor{firstColor}{27.576} & \cellcolor{firstColor}{0.952} & \cellcolor{firstColor}{0.041} \\
PuzzleAvatar   & \cellcolor{secondColor}24.687 & \cellcolor{secondColor}0.930 & \cellcolor{secondColor}0.062 \\
TECH           & \cellcolor{thirdColor}23.635 & \cellcolor{thirdColor}0.919 & \cellcolor{thirdColor}0.065 \\
AvatarBooth    & 16.431 & 0.758 & 0.153 \\
\hline
\end{tabular}
\caption{\textbf{Quantitative comparison of PuzzleIOI dataset.} Compared to existing baselines, PuzzleAvatar, TECH, and Avatarbooth. PFAvatar achieves the best performance across all evaluation metrics.}
\label{tab:compare_psnr_metric}
\end{table}

\begin{table}[H]
\centering
    \scalebox{0.65}{
    \begin{tabular}{lcccccc}
        \toprule
        \multirow{2}{*}{\centering Method} & \multicolumn{2}{c}{CLIP-I $\uparrow$} & \multicolumn{2}{c}{DINO $\uparrow$} & \multicolumn{2}{c}{CLIP-T $\uparrow$} \\
        \cmidrule(lr){2-3} \cmidrule(lr){4-5} \cmidrule(lr){6-7}
        & full-body & head & full-body & head & full-body & head \\
        \midrule
        Full & \cellcolor{firstColor}0.9125 & \cellcolor{firstColor}0.9042 & \cellcolor{firstColor}0.8072 & \cellcolor{firstColor}0.8617 & \cellcolor{firstColor}0.3546 & \cellcolor{firstColor}0.3124 \\
        w/o Head Part Data &  \cellcolor{thirdColor}0.8702 & 0.8503 & 0.7154 & 0.8032 & \cellcolor{thirdColor}0.2912 & \cellcolor{thirdColor}0.2538 \\
        w/o ControlBooth & 0.8352 & 0.8234 & 0.7091 & 0.8023 & 0.2314 & 0.2313 \\
        w/o 3D-SDS & 0.8021 & 0.8127 & 0.7281 & 0.7819 & 0.2281 & 0.2134 \\
        w/o $\mathcal{L}_{\text{geo}}$ & \cellcolor{secondColor}0.8929 & \cellcolor{secondColor}0.8831 & \cellcolor{secondColor}0.8011 & \cellcolor{secondColor}0.8590 & \cellcolor{secondColor}0.3257 & \cellcolor{secondColor}0.3081 \\
        w/o Multi-sampling & 0.8654 & \cellcolor{thirdColor}0.8612 & \cellcolor{thirdColor}0.7486 & \cellcolor{thirdColor}0.8574 & 0.2812 & 0.2434 \\
        \bottomrule
    \end{tabular}}
    \caption{\textbf{Quantitative comparison of ablation results.} Subject fidelity (DINO, CLIP-I) and prompt fidelity (CLIP-T, CLIP-T-L) scores under different ablation settings.}
    \label{tab:ablation-study}
\end{table}

\section{Conculusion}

In this paper, we introduce PFAvatar, a novel method for reconstructing high-quality 3D avatars from real-world “Outfit of the Day”(OOTD) photos. Our approach consists of two key stages: (1) fine-tuning a Pose-Aware Diffusion Model using few-shot OOTD examples, and (2) distilling a 3D NeRF-based avatar. By avoiding decomposition and directly modeling full-body appearance, integrating ControlNet for pose estimation, and introducing a Condition Prior Preservation Loss (CPPL), our method achieves consistent and detailed results while mitigating language drift. Additionally, our NeRF-based representation, optimized through canonical SMPL-X space sampling and Multi-Resolution 3D-SDS, effectively handles occlusions and preserves high-frequency textures. Experiments demonstrate that PFAvatar outperforms state-of-the-art methods in reconstruction fidelity, detail preservation, and robustness to occlusions/truncations, significantly advancing the practical generation of 3D avatars from real-world OOTD albums. In addition, we demonstrate the downstream applicability of the reconstructed 3D avatars, including virtual try-on, animation, and human video reenactment.
\section*{Acknowledgments}
The work was partially supported by National Key R\&D Program of China (No. 2023YFF0905102), Key R\&D Program of Zhejiang Province (No. 2023C01039)
\bibliography{aaai2026}
\end{document}